\documentclass[english]{article}
\usepackage[T1]{fontenc}
\usepackage[latin9]{inputenc}
\usepackage{babel}
\usepackage{array}
\usepackage{multirow}
\usepackage{graphicx}
\usepackage{hyperref}
\usepackage{wrapfig}
\usepackage[table]{xcolor}
\usepackage{booktabs}
\usepackage{amsfonts}
\usepackage{nicefrac}
\usepackage{microtype}
\usepackage{authblk}
\usepackage{float}
\makeatother

\begin{document}

\title{Self-paced Convolutional Neural Network for Computer Aided Detection in Medical Imaging Analysis}

\author[1]{Xiang Li}
\author[2]{Aoxiao Zhong}
\author[4]{Ming Lin}
\author[1]{Ning Guo}
\author[3]{Mu Sun}
\author[4]{Arkadiusz Sitek}
\author[4]{Jieping Ye}
\author[1]{James Thrall}
\author[1]{Quanzheng Li}
\affil[1]{Massachusetts General Hospital}
\affil[2]{Zhejiang University}
\affil[3]{Beijing Institute of Technology}
\affil[4]{University of Michigan}

\maketitle

\begin{abstract}
Tissue characterization has long been an important component of Computer Aided Diagnosis (CAD) systems for automatic lesion detection and further clinical planning. Motivated by the superior performance of deep learning methods on various computer vision problems, there has been increasing work applying deep learning to medical image analysis. However, the development of a robust and reliable deep learning model for computer-aided diagnosis is still highly challenging due to the combination of the high heterogeneity in the medical images and the relative lack of training samples. Specifically, annotation and labeling of the medical images is much more expensive and time-consuming than other applications and often involves manual labor from multiple domain experts. In this work, we propose a multi-stage, self-paced learning framework utilizing a convolutional neural network (CNN) to classify Computed Tomography (CT) image patches. The key contribution of this approach is that we augment the size of training samples by refining the unlabeled instances with a self-paced learning CNN. By implementing the framework on high performance computing servers including the NVIDIA DGX1 machine, we obtained the experimental result, showing that the self-pace boosted network consistently outperformed the original network even with very scarce manual labels. The performance gain indicates that applications with limited training samples such as medical image analysis can benefit from using the proposed framework.
\end{abstract}
 
\section{Introduction}

In medical image analysis, tissue characterization and classification are among the most important components in a Computer Aided Diagnosis (CAD) system. An accurate and robust tissue classifier is one of the ultimate goals for many radiology applications. In recent years, impressive improvements on various computer vision problems have been reported using deep learning-based models over traditional machine learning and statistical methods. Specifically, Convolutional Neural Networks (CNN) have shown superior capabilities in extracting the low to high-level image features needed to perform the classification with the deep neural networks. These successes have motivated the increasing application of deep learning for medical image analysis \cite{RN546,MitosisCiresan2013,RN679,RN691}. However, it has also been recognized that deep learning models (and actually most of the learning-based methods) are far more difficult to be successfully applied on medical image analysis comparing with the natural image analysis. One of the main challenges arises from the limited number of labeled samples for training the model \cite{RN567}, as annotation and labeling of medical images is a highly time consuming and labor-intensive work. Also, medical images are highly heterogeneous both on an individual-level and population-level. The combination effect of these two limitations severely degrades the robustness of the models and the reproducibility of the learning results. The highly sensitive nature of the medical images regarding privacy concerns poses difficulty in cross-institutional data sharing which further limits the availability of case material. 

On the other hand, the availability and size of medical images on public domains have been increasing very fast over the past few years. However most of them are not annotated as these databases are usually provided for general purpose of use, and annotation is extremely costly. Consequently, there exist huge discrepancies between the large number of datasets to be analyzed and the very limited number of available annotations to be used as training data. In response to the challenge of the lack of training samples, we propose the self-paced Convolutional Neural Network (spCNN) framework which is able to identify unlabeled image patches as "virtual" training samples. These virtual samples are then mixed with the original manually-labeled samples to retrain a new network. By introducing the virtual samples, we can practically obtain any number of training samples by increasing the computational load of the machines in exchange for the (much more expensive) human labor work, provided with sufficient unlabeled images to be analyzed. As the database used in this work is constitutes the imaging data from 10,000 subjects, the potential number of virtual samples can be huge to support highly complicated learning.

We tested the performance of both the raw CNN trained only from manually-labeled samples and the CNN retrained on the mixed training samples, by applying them both to another benchmark testing dataset labeled by a different group of experts. The classification results show that the accuracy of the spCNN framework is improved by over 10\% with the help of the virtual samples. The improvement is shown to be consistent using different network architectures and parameter settings. The performance improvement show that the proposed spCNN framework could provide a new perspective on improving the performance of learning-based methods in medical image analysis as well as other applications with limited training samples. That is, we can leverage small-size training data to train an initial network, then adaptively select the virtual samples to add them back into the training set to improve the performance in a "snow-balling" fashion.
 
\section{Related Works}

As the name implies, our spCNN framework is motivated by self-paced learning \cite{Jiang:2014:SLD:2969033.2969059,Kumar:2010:SLL:2997189.2997322,Jiang:2015:SCL:2886521.2886696} and curriculum learning \cite{Bengio:2009:CL:1553374.1553380} methods, which adaptively choose part of labeled instances for the training. Instances considered to be easy are selected first, such as those with large margin or high confidences. Difficult instances will be learned in later stages or even dropped eventually. Most of the self-paced learning methods focus on identifying the optimized order of learning and learning the model simultaneously \cite{SupancicIII:2013:SLL:2514950.2516284,}. In our work CNNs are trained on an initial set of data followed by a bootstrapping scheme to evaluate the unlabeled instances. Then a new CNN is trained based on the selected instances.
Another scheme related with our framework is learning from positive and unlabeled examples (PU learning) \cite{Elkan:2008:LCO:1401890.1401920}, where only positive instances are partially labeled. Thus, it is needed in PU learning to identify a set of reliable negative-labeled instances from the unlabeled data and use them for the further training \cite{Li:2010:NTD:1870658.1870680}, which is similar to our problem yet more focusing on the estimation of the labels without utilizing the learning-based method as we do. 

It is also worth noting that several semi-supervised learning techniques including ladder networks \cite{Rasmus:2015:SLL:2969442.2969635} which incorporate auto-encoder into the supervised model with skipped connections, as well as the stacked what-where auto-encoders \cite{Vincent:2010:SDA:1756006.1953039} that utilize both convolution and deconvolution nets to allow integrated supervised and unsupervised learning. While our spCNN model does not incorporate the unsupervised component, it is advantageous over the semi-supervised learning approaches in that we can obtain an explicit evaluation of the new data and select the samples accordingly based on simple rules, which could be highly useful for clinical practice and decisions.

From an application perspective, there are various literature reports that applied CNN models to analyze medical images with lung diseases \cite{RN689,RN690,Gao2016,Shen2015} and obtained encouraging results. As we discussed previously, the lack of training data is especially challenging in medical image analysis. Current solutions for addressing this issue \cite{RN679} include traditional data augmentation techniques based on affine transformations to expand the labeled dataset \cite{Shen2015,7279156,7422783}. Also, researchers have tried utilizing the vast amount of labeled natural images to help training the network in a learning fashion \cite{RN691,RN694} or initialize network parameters (i.e. weights) by pre-training the network with non-medical images then fine-tuning the parameters with the target samples \cite{Gupta:2013:NIB:3042817.3043047,Brosch2013}. Although we are focusing on unlabeled medical data, which is conceptually different from these works that focus on the labeled non-medical data. These two approaches are not mutually exclusive however, and can be combined together for an integrated and more effective framework.
 
\section{Materials and Methods}

\subsection{Method overview}

In this work, we propose the self-paced Convolution Neural Network (spCNN) framework in order to improve the accuracy and robustness of the learned model beyond the information provided from the initially limited training data. The major contribution and novelty of the proposed framework is that it leverages the large amount of new, unlabeled data as potential new training samples for the retraining. The new training samples are called "virtual samples" in contrast to the original manually-labeled samples. Specifically, class labels and the distribution of prediction accuracies of the samples in the new dataset are estimated by bootstrapping CNNs. Image patches with significant different prediction probabilities across labels are then pooled together with the original training data for retraining a new CNN. A conceptual diagram of the framework design is illustrated in Fig.1. 

\begin{figure}[H]
	\begin{centering}
		\includegraphics[width=0.7\columnwidth]{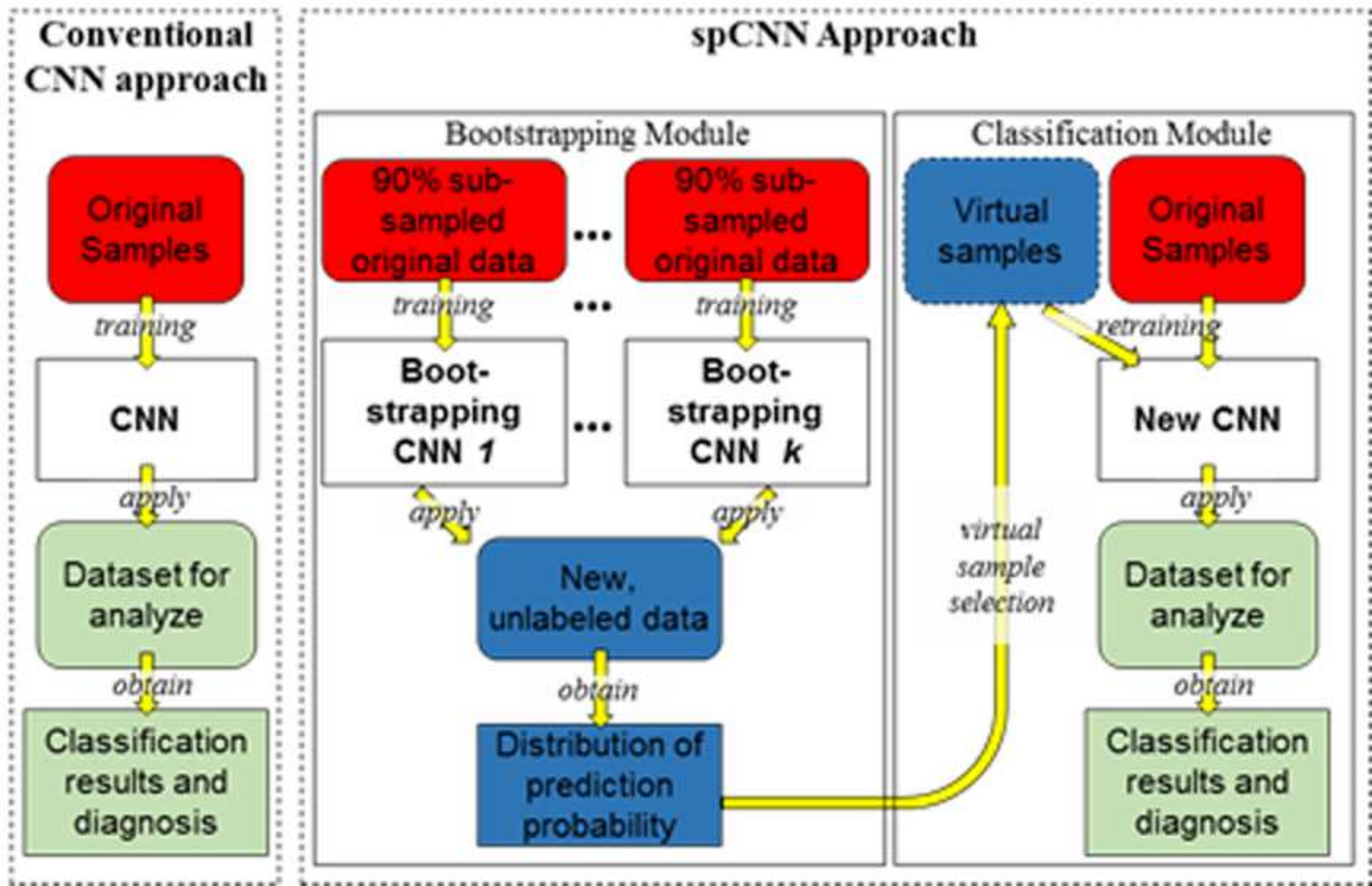}
		\caption{Left: Illustration of the conventional CNN approach of performing image classification. Right: Illustration of the spCNN framework consists of the main ?Classification Module? similar to the conventional approach, and the ?Bootstrapping Module? which provides the extra training data through virtual sample selection based on the bootstrapping CNNs performed on the new unlabeled data.}
	\end{centering}
\end{figure} 

\subsection{Architecture of the CNN applied in the framework}

 The CNN model used in the proposed framework is implemented in Caffe \cite{Jia:2014:CCA:2647868.2654889}, and its architecture is shown in Fig. 2. Image patches of size 36$×$36 are convolved by 4 convolutional layers. The kernel size of all convolutional layers is set to 3. Based on the principles introduced in \cite{RN546} that the number of kernels in each layer shall be proportional to the area of its receptive field (in this work, from 3$×$3 in the first layer to 6$×$6 in the fourth layer), we set the number of kernels in the four layers as 45, 80, 125 and 180 respectively. Each convolutional layer is followed by a maximum pooling layer. The extracted features are then fed into three fully connected layers, with the number of neurons being 1080, 360 and 3. These numbers are proportional (6 and 2 times) to the number of features (180), based on the empirical rules reported in \cite{RN546}. The first two fully connected layers are equipped with dropout layers \cite{JMLR:v15:srivastava14a} with probability of 50\%. Both the convolutional layers and the fully connected layers use the activation function of LeakyReLU \cite{reLU_ICML}. 
\begin{figure}[H]
	\begin{centering}
  		\includegraphics[width=0.7\textwidth,height=0.088\textwidth]{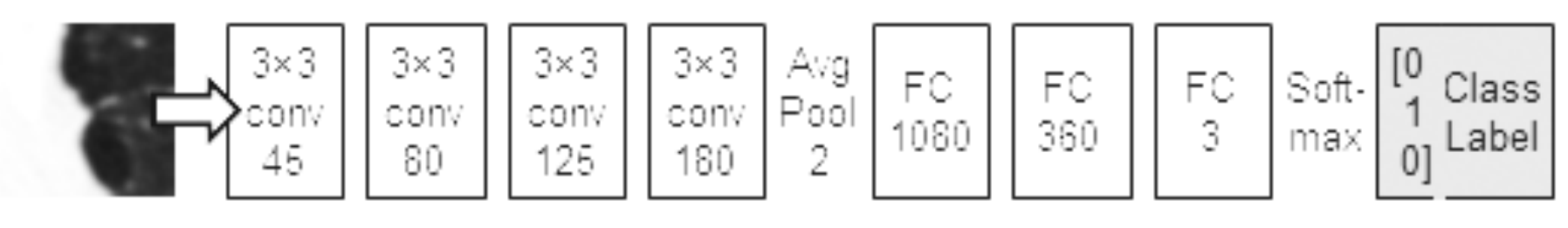}
  		\caption{Architecture of the CNN used in the proposed framework. Each convolutional layer and fully-connected layer is followed by a LeakyReLU activation function.}
  	\end{centering}
\end{figure}

\subsection{Bootstrapping module for virtual sample selection}
The key challenge of the spCNN framework is how to correctly select the image patches from the new dataset into the virtual samples: labels of the patches could be wrongfully assigned by the initially trained CNN which is clearly not desired. Specifically, it has been observed in both our experiments and in previous work \cite{Jiang:2015:SCL:2886521.2886696} that the errors of the model could be quickly accumulated during the retraining and eventually lead to a performance decrease. On the other hand, we will also want to push the boundary of the retraining dataset beyond the original manual annotations, which often involves image patches with higher uncertainty of the accuracy from the network. In other words, the framework needs to balance between the original manually-labeled samples and the latterly identified virtual samples through the learning process.

Thus, in this work we apply a 10-folds bootstrapping scheme to estimate the empirical distribution of the predication probabilities of the new samples, and select the most suitable samples automatically according to the statistical testing. Specifically, we perform random subsampling for 10 times to obtain 10 sets of 90\% of the original training data, which are 360 patches for each class for a total of 1080 patches. These 10 sets of data are then used to train the bootstrapping networks. Patches in the new dataset are then classified by each of the 10 networks, resulting in 10 sets of class labels and prediction probabilities for each patch, illustrated in Fig. 3.

\begin{figure}[H]
	\begin{centering}
		\includegraphics[width=0.6\columnwidth]{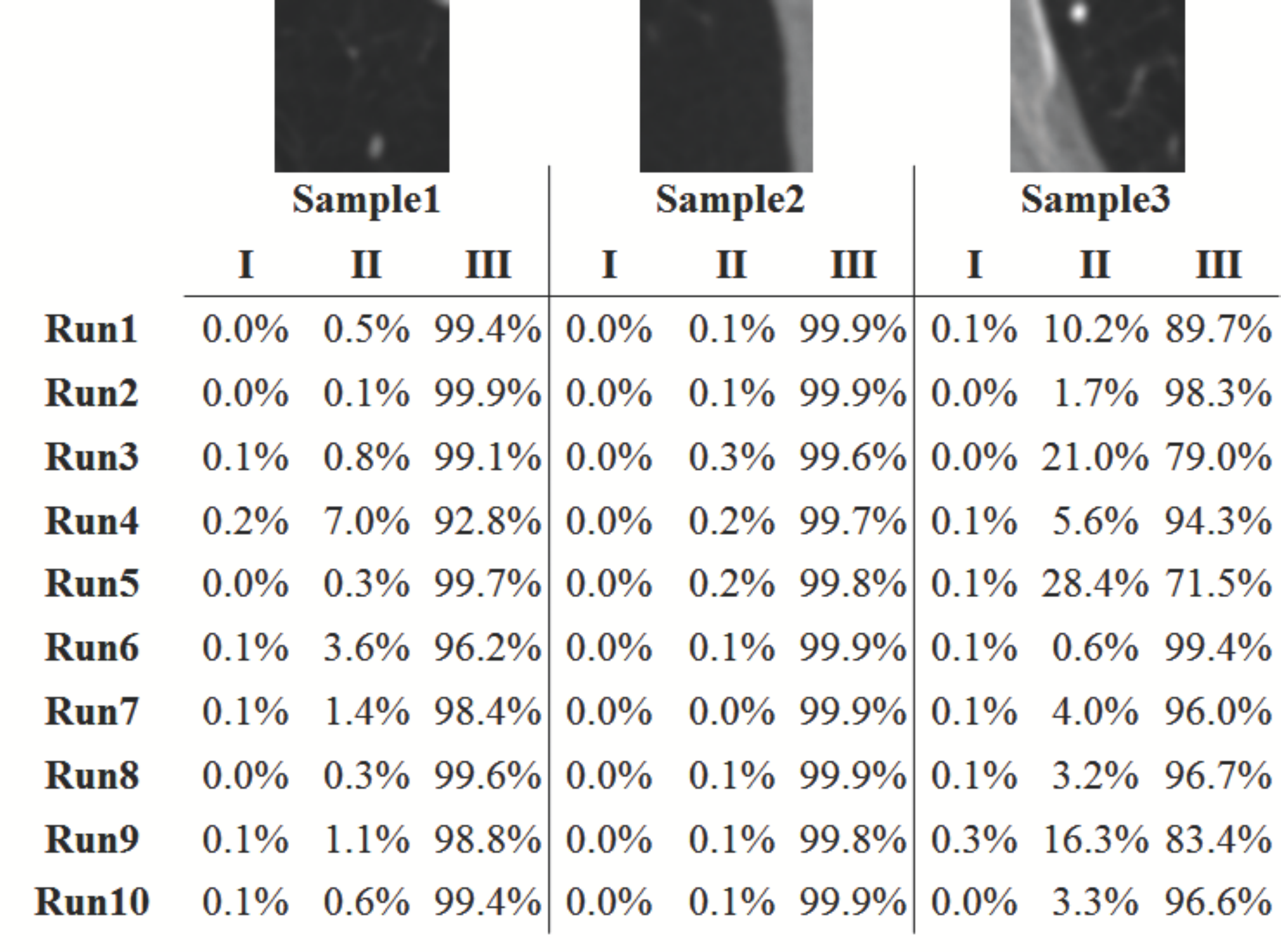}
		\caption{Illustration of the 10-folds bootstrapping results of the new data on 3 example image patches. Captions of the table columns (I, II and III) indicate the prediction probability of the given patch belonging to the corresponding class.}
	\end{centering}
\end{figure}

It could be found that while certain patches (e.g. the first two patches in Fig.3) in the new dataset can be easily classified with high prediction probabilities and low variability from all the 10 bootstrapping networks, there are cases where the classification uncertainty is much higher (e.g. the third patch). However, visual inspection indicates that the third image patch lies on the boundary between the normal lung tissue and regions outside of lung, which is definitely a good candidate of a virtual sample to be included in the retraining process due to the fact that: 1) the bootstrapping networks unanimously assign a correct label to it, and 2) such patches are much rarer in the manual annotation results and underrepresented in the training set. As reported in \cite{doWeNeedMoreTrainingData}, the sheer number of training samples does not help too much for training the network especially if they come from a homogeneous population. It is the samples that are not encountered before will actually lead to better performance. Based on such observations, we also deduce that virtual samples can help expanding the solution space the training process can explore thus improve it. 

Thus, in this work, for each of the 10$×$3 prediction probability matrix of the given image patch estimated by the bootstrapping CNNs, we will perform two two-sample t-tests (as there are totally 3 labels), aiming to find whether the label with highest average prediction probability is significantly higher than both of the other two labels. The two p-values produced by the t-tests from each patch will be then aggregated and further analyzed by the false discovery rate (FDR) control respectively. Here we employed the FDR to minimize the possibility that the huge number of testing performed could lead to increased false positives (i.e. unfitting patches). Patches with significantly different prediction probabilities across the 3 labels will be selected as virtual samples for retraining the new CNN.  

\section{Experimental Results}

\subsection{Data acquisition and preprocessing}

In this work, we use the data from the COPDGene database \cite{doi:10.1056/NEJMoa1007285} sponsored by NIH, which aims to investigate the CT phenotypes in Chronic Obstructive Pulmonary Disease (COPD) and other lung diseases. For the purpose of testing and validating the proposed model, we mainly focus on pulmonary emphysema, defined as the permanent enlargement of airspaces distal to the terminal bronchioles and the destruction of the alveolar walls. In the COPDGene database, 3-D volumetric images are acquired using 64-slice CT scanners during full inspiration, and then reconstructed using sub-millimeter slice thickness with smoothing and edge-enhancing filters. From the total of 10,000 subjects in the database from both normal and COPD population, 500 image slices from 150 subjects were manually annotated by a group of experts on our team for the three classes: airway (Class I), emphysema (Class II), and other lung tissue (Class III). For each of the 3 classes, 600 non-overlapping image patches of size 36$×$36 were extracted from the annotation results, constituting the samples for training (400 patches) and verification (200 patches). An example set of the image patches from the three classes is shown in Fig. 4.

\begin{figure}[H]
	\begin{centering}
		\includegraphics[width=0.6\columnwidth]{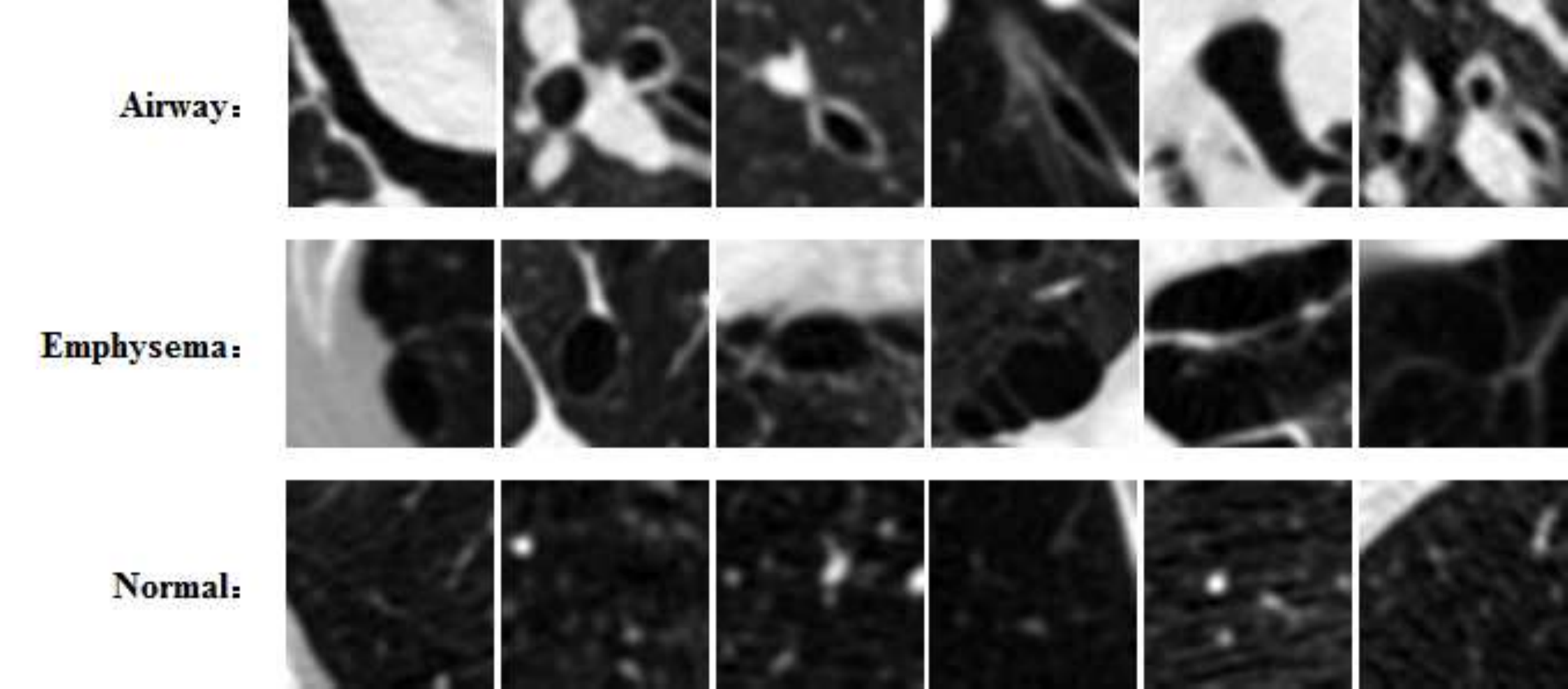}
		\caption{Six sample image patches from each of the 3 classes.}
	\end{centering}
\end{figure}

At the same time, from the new unlabeled dataset, 9600 patches were extracted and analyzed by the bootstrapping CNNs, constituting the candidates for virtual sample selection. Finally, 161 image slices from another 59 subjects were manually annotated by another group of experts. Totally 887 (Class I: 203, Class II: 192, Class III: 255) image patches were extracted from the annotated regions, which were used as the benchmark testing inputs for evaluating the model performance.

\subsection{Performance comparisons}

By applying the bootstrapping module of the proposed spCNN framework on the 9600 image patches in the new dataset, we select the virtual samples according to the different significant level. We then retrain the new CNN models from the mixture of the virtual samples and the original manually-labeled samples. The new CNNs are then applied to classify the benchmark testing dataset. The model performance and the details of the virtual samples are summarized in Table 1. The results show that using the significant level $p$=0.05/0.1 for the FDR-controlled statistical testing, spCNN can obtain as high as 10\% of accuracy increase over the raw CNN model trained solely from the original manually-labeled samples. The classification accuracy which is near 90\% is on the same level with the results from a similar lung CT image study using CNN as reported in \cite{RN691}. Using a more conservative significant level ($p$=0.025), fewer data will be selected, obtain similar levels of accuracy compared to the raw CNN without causing performance decrease. 

\begin{table}[ht]
\caption{Comparison of the number of virtual samples selected (out of 9600 patches) as well as the classification accuracies among the raw CNN (first row) and spCNN under different significant levels for FDR-controlled statistical testing.}
\vskip 0.1in
\label{tableAccuracyComparison}
\begin{center} 
\begin{small}
\begin{sc}
\begin{tabular}{ccc}
\hline
%\abovespace
%\belowspace
Model&Number of virtual samples&Accuracy\\
\hline
%\abovespace
Original samples only&N/A&79.0\%\\  
$p$=0.025&665&84.1\%\\  
$p$=0.05&891&87.3\%\\  
$p$=0.1&943&88.9\%\\  
\hline
\end{tabular}
\end{sc}
\end{small}
\end{center}
\vskip -0.1in
\end{table}

As the current virtual sample selection in spCNN are empirically determined using bootstrapping scheme, one important question is whether the framework are robust enough to guide the virtual sample selection process under different models and/or different datasets. Limited by the size and scope of this manuscript, we only focused on the COPDGene dataset, yet test the spCNN performance using the same virtual sample selection method but different CNN architectures. Specifically, we have tried replacing the CNN architecture as introduced in 2.1 by the following designs: 1) Reducing the number of kernels in the convolutional layers as well as the number of neurons in the fully connected layers by half. 2) Increasing the number of kernels in the convolutional layers as well as the number of neurons in the fully connected layers by 50\%. 3) Adding an extra fully connected layer with number of neurons of 180 before the last layer. 4) Removing the first 3$×$3 convolutional layer. The results show that, while the classification performance of the spCNN framework based on these 4 network architectures varies, in all of the cases using significant level of $p$=0.1 will outperform other configurations, as well as the original CNN method.

\subsection{Time cost for the virtual sample selection}
As we previously discussed, the self-paced learning framework essentially exchanges human labor work with computational costs through the bootstrapping process. So, the time cost for training the bootstrapping CNNs could be an important factor for the proposed spCNN framework, especially for larger datasets and/or more complicated network architectures. Currently we deploy the framework on two platforms: one is an in-house server installed with two NVIDIA Tesla P-100 GPUs. The other is the NVIDIA DGX-1 deep learning system with eight P-100 GPUs interconnected with the NVIDIA NVLink \cite{RN681}. Time costs for training one bootstrapping CNN in the bootstrapping module, as well as for performing one forward-backward propagation using different hardware configurations are listed in Table 2. 

\begin{table}[ht]
\caption{Time cost for training one CNN during bootstrapping ("per session", measured in seconds) and time cost for performing one forward-backward propagation ("per iteration", measured in milliseconds). All values are estimated by averaging from the time costs of running the CNN training for 100 times.}
\vskip 0.1in
\label{tableSpeedComparison}
\begin{center} 
\begin{small}
\begin{sc}
\begin{tabular}{ccc}
\hline
%\abovespace\belowspace
Configuration&per Session (s)&per Iteration (ms)\\
\hline
%\abovespace
In-house, single GPU&50.63&5.01\%\\  
In-house, 2 GPUs&50.92&5.00\%\\  
DGX1, single GPU&44.46&4.41\%\\  
DGX1, 2 GPUs&33.80&3.32\%\\  
DGX1, 4 GPUs&27.84&2.66\%\\  
DGX1, 8 GPUs&27.98&2.54\%\\  
\hline
\end{tabular}
\end{sc}
\end{small}
\end{center}
\vskip -0.1in
\end{table}

It can be seen that using the most advanced accelerator of NVIDIA DGX1, we can achieve a nearly 2-fold speed increase. It should be noted that the in-house sever shows a lowered running speed using two GPUs comparing with a single GPU, due to the fact that P2P DMA access between devices is needed for running Caffe in multiple-GPU mode. When the P2P access is not supported (as in our in-house server), data will need to be copied through hosts thus severely affect the performance. On the contrary, the DGX1 system is much better optimized for parallelizing the computational loads across multiple GPUs, showing the importance of the P2P DMA access and the NVIDIA NVLink technology. Also, we observe that using 8 GPUs in DGX1 does not result in better performance comparing with 4 GPUs even though the iteration time has been reduced, which is most likely due to the fact that the overhead for parallelization became dominant in the time cost for 8 GPUs. Considering the fact that the current CNN architecture is relatively simple, we envision that 8 GPUs will outperform other configurations in more complicated cases.
 
\section{Conclusion and Discussion}
In this work, we develop the self-paced scheme for identifying virtual samples from the unlabeled data and use them to retrain a new CNN, in order to overcome the problems of the lack of training samples. The FDR-controlled statistical testing for the virtual sample selection based on bootstrapping scheme shows that the current optimized threshold is around $p$=0.1. Similar threshold could be used for analyzing the rest of the data within the COPDGene dataset as such inference is not affected by the number of classes nor the number of samples tested. Actually, we propose that the parameter tuning on the threshold is essentially the empirical characterization for the relationship between the distribution of the network outputs and the quality/confidence of the corresponding samples which is highly related with the dataset. According our experience in the self-paced learning procedure, we envision that further improve for the performance of spCNN could be achieved by gradually selecting the virtual samples in multiple rounds, similar to the majority of curriculum learning methods. In other words, after selecting the virtual samples and retraining the new CNN, we can update the bootstrapping CNNs based on the mixture data (with subsampling), then perform the same virtual sample selection again on the remaining dataset. The multiple-rounds application of spCNN thus form a closed loop from the selected virtual samples back to the selection process. 
 
\bibliographystyle{unsrt}
\bibliography{manuscript_spCNN}

\end{document}